\documentclass[sigconf]{acmart}

\AtBeginDocument{%
  \providecommand\BibTeX{{%
    \normalfont B\kern-0.5em{\scshape i\kern-0.25em b}\kern-0.8em\TeX}}}

\setcopyright{acmcopyright}
\copyrightyear{2024}
\acmYear{2024}
\acmDOI{XXXXXXX.XXXXXXX}

\acmISBN{978-1-4503-XXXX-X/18/06}

\usepackage{xspace}
\makeatletter
\DeclareRobustCommand\onedot{\futurelet\@let@token\@onedot}
\def\@onedot{\ifx\@let@token.\else.\null\fi\xspace}

\def\ie{\emph{i.e}\onedot} 
 
\def\etc{\emph{etc}\onedot} 
 
\def\etal{\emph{et al}\onedot}
\makeatother

\definecolor{refblue}{rgb}{0.21,0.49,0.74}
\makeatletter
\@ifpackageloaded{hyperref}{
}{
    \usepackage[pagebackref,breaklinks,colorlinks,citecolor=refblue]{hyperref}
}
\makeatother

\usepackage[group-separator={,}]{siunitx}
\usepackage{amsmath}
\usepackage{amsfonts}
\usepackage{makecell}
\usepackage{lipsum}
\usepackage{tcolorbox}
\usepackage{wrapfig}

\usepackage{xcolor}
\usepackage{colortbl}
\usepackage{multirow}
\usepackage{tabularx}
\usepackage{booktabs} 
\usepackage{graphbox}
\usepackage{footmisc}
\usepackage{bm}
\usepackage{caption}
\usepackage{arydshln}
\usepackage{dashrule}
\usepackage{enumitem}
\setlist[itemize]{noitemsep,nolistsep}
\usepackage{tcolorbox}
\usepackage{multicol} 
\usepackage{fancyvrb}
\usepackage{listings}
\usepackage{setspace}

\usepackage{algorithm}
\usepackage{algorithmic}
\usepackage[labelformat=simple]{subcaption}

\usepackage{tablefootnote}

\usepackage[capitalize]{cleveref}
\crefname{section}{Sec.}{Secs.}
\Crefname{section}{Section}{Sections}
\Crefname{table}{Table}{Tables}
\crefname{table}{Tab.}{Tabs.}
\Crefname{figure}{Figure}{Figures}
\crefname{figure}{Fig.}{Figs.}
\Crefname{equation}{Equation}{Equations}
\crefname{equation}{Eq.}{Eqs.}
\Crefname{algorithm}{Algorithm}{Algorithms}
\crefname{algorithm}{Alg.}{Algs.}

\usepackage[labelformat=simple]{subcaption}

\colorlet{lightpink}{pink!35}
\colorlet{lightcyan}{cyan!20}
\colorlet{red}{red!80}
\colorlet{blue}{blue!80}
\colorlet{green}{green!60!black}
\colorlet{algemp}{cyan!10}
\colorlet{lightgray}{gray!20}

\colorlet{jitter}{blue!60}
\definecolor{noise}{HTML}{93FFF0}
\definecolor{overexposure}{HTML}{CDE55D}
\definecolor{blur}{HTML}{E45C5C}
\definecolor{lowlight}{HTML}{35E344}

\usepackage{dirtree}
\usepackage{xcolor,colortbl}

\usepackage{pifont}
\newcommand{\cmark}{\textcolor{green}{\ding{52}}}%
\newcommand{\xmark}{\textcolor{red}{\ding{56}}}%

\newcolumntype{C}[1]{>{\centering\arraybackslash}p{#1}}
\newcolumntype{L}[1]{>{\raggedleft\arraybackslash}p{#1}}
\newcolumntype{R}[1]{>{\raggedright\arraybackslash}p{#1}}
\newcolumntype{a}{>{\columncolor{gray!20!white}}c}

\graphicspath{{./figures/}{./supp_figs/}}

\newcommand{\dataset}{QGround-100K }
\newcommand{\prompt}[1]{``{\fontfamily{cmtt}\selectfont #1}''}

\usepackage{tikz}
\newcommand{\dottedbox}[2][]{%
  \begin{tikzpicture}
    \node[draw, dotted, thick, minimum width=\columnwidth, inner sep=2pt, anchor=text, rectangle, #1] {
    \begin{minipage}{.95\linewidth}#2\end{minipage}};
  \end{tikzpicture}
}

\begin{document}

\title{Q-Ground: Image Quality Grounding with Large \\ Multi-modality Models}

\settopmatter{authorsperrow=4}
\author{Chaofeng Chen}
\affiliation{%
  \institution{S-Lab, Nanyang Technological University}
  \city{}
  \country{}}
\email{chaofenghust@gmail.com}

\author{Sensen Yang}
\affiliation{%
  \institution{S-Lab, Nanyang Technological University}
  \city{}
  \country{}}
\email{s230059@e.ntu.edu.sg}

\author{Haoning Wu}
\affiliation{%
  \institution{S-Lab, Nanyang Technological University}
  \city{}
  \country{}}
\email{haoning001@e.ntu.edu.sg}

\author{Liang Liao}
\affiliation{%
  \institution{S-Lab, Nanyang Technological University}
  \city{}
  \country{}}
\email{liang.liao@ntu.edu.sg}

\author{Zicheng Zhang}
\affiliation{%
  \institution{Shanghai Jiao Tong University}
  \city{}
  \country{}}
\email{zzc1998@sjtu.edu.cn}

\author{Annan Wang}
\affiliation{%
  \institution{Nanyang Technological University}
  \city{}
  \country{}}
\email{c190190@e.ntu.edu.sg}

\author{Wenxiu Sun \\ Qiong Yan}
\affiliation{%
  \institution{SenseTime Research}
  \city{}
  \country{}}
\email{[sunwx,yanqiong]@tetras.ai}


\author{Weisi Lin}
\affiliation{%
  \institution{Nanyang Technological University}
  \city{}
  \country{}}
\email{wslin@ntu.edu.sg}
\renewcommand{\shortauthors}{Chaofeng Chen et al.}

\begin{abstract}

Recent advances of large multi-modality models (LMM) have greatly improved the ability of image quality assessment (IQA) method to evaluate and explain the quality of visual content.
However, these advancements are mostly focused on overall quality assessment, and the detailed examination of local quality, which is crucial for comprehensive visual understanding, is still largely unexplored.
In this work, we introduce \textbf{Q-Ground}, the first framework aimed at tackling fine-scale visual quality grounding by combining large multi-modality models with detailed visual quality analysis.
Central to our contribution is the introduction of the \textbf{QGround-100K} dataset, a novel resource containing 100k triplets of \textit{(image, quality text, distortion segmentation)} to facilitate deep investigations into visual quality. 
The dataset comprises two parts: one with human-labeled annotations for accurate quality assessment, and another labeled automatically by LMMs such as GPT4V, which helps improve the robustness of model training while also reducing the costs of data collection.
With the \textbf{QGround-100K} dataset, we propose a LMM-based method equipped with multi-scale feature learning to learn models capable of performing both image quality answering and distortion segmentation based on text prompts. This dual-capability approach not only refines the model's understanding of region-aware image quality but also enables it to interactively respond to complex, text-based queries about image quality and specific distortions. 
\textbf{Q-Ground} takes a step towards sophisticated visual quality analysis in a finer scale, establishing a new benchmark for future research in the area. Codes and dataset are available at \url{https://github.com/Q-Future/Q-Ground}.

\end{abstract}

\ccsdesc[100]{Computing methodologies~Computer vision tasks}

\keywords{Image Quality Assessment, Large Multi-modality Models, Image Quality Grounding}

\maketitle

\section{Introduction} \label{sec:intro}

\begin{figure}[!t]
    \centering
    \includegraphics[width=\linewidth]{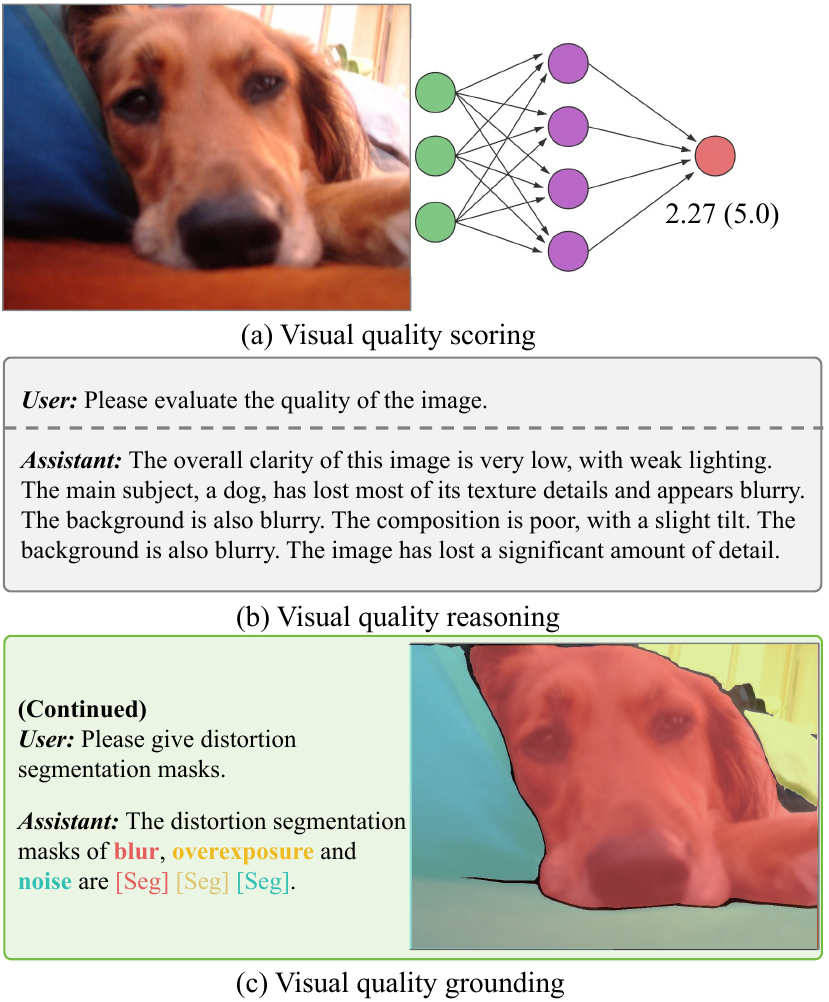}
    \caption{An example comparison between different tasks illustrates: (a) Visual quality scoring only provides a numerical score without an underlying rationale; (b) LMM-based reasoning offers clear explanations but lacks pixel-level comprehension; (c) the suggested approach to visual quality understanding not only facilitates quality reasoning but also delivers corresponding pixel-level distortion segmentation masks.}
    \label{fig:intro}
\end{figure}

As a cornerstone in the domain of digital imaging, Image Quality Assessment (IQA) covers a wide range of methods designed to evaluate the perceptual quality of visual content similarly to human visual system (HVS). With the rapid increase of digital content, IQA is becoming more and more important in many areas, such as media streaming, user-generated photos and videos, smart-phone cameras and the growing field of AI-generated content. These various applications call for more powerful and understandable IQA methods to help create visual content with better quality and improve the experience of users. 

Existing IQA methods has aimed to replicate the HVS's capability to distinguish and assess visual information, typically by correlating the mean opinion scores (MOS) labeled by humans with features derived from images. The performance of these methods has significantly improved with the advent of more powerful feature extractors, moving from hand-crafted features in traditional approaches 
 \cite{niqe,ilniqe,brisque,ma2017nrqm} to advanced deep neural networks \cite{zhang2018lpips,musiq,topiq}. 
Nonetheless, these works only give quality scores as results and face challenges in accurately evaluating and explaining the details of image quality, particularly when it comes to local distortions and fine-grained analysis. 
Recent advances in Large Multi-Modality Models (LMMs) mark a new chapter for IQA, offering promising avenues for enhancing both the evaluation capabilities and the explanatory ability of IQA systems. 
For example, CLIPIQA \cite{clipiqa} illustrates the zero-shot capabilities of multi-modality models in IQA, and Q-Bench \cite{wu2023qbench} demonstrates near-human performance of GPT4V \cite{gpt4v} in certain specific areas. However, despite these advancements, the application of LMMs in IQA remains focused on overall quality assessment. This narrow focus limits their utility for comprehensive visual analysis, particularly in contexts where fine-scale quality grounding and detailed understanding of local distortions are imperative. 

In response to these challenges, we introduce the visual quality grounding task to the field of IQA for the first time, with the goal of bridging the gap in detailed image quality perception. 
As illustrated in \cref{fig:intro}, traditional methods of quality scoring yield a single numerical score without explanation, and existing quality reasoning methods does not account for local distortions. 
Our novel visual quality grounding strategy integrates pixel-level distortion segmentation with textual queries, substantially improving the fine-scale capabilities of IQA. 
The major problem in realizing this advancement is the lack of suitable datasets.
Unlike standard segmentation tasks, the boundaries of distortion regions may exhibit minor variations due to individual subjective judgments. 
Therefore, we deploy two auxiliary methods to support the creation of more dependable mask annotations: 1) Preliminary segmentation of images using Semantic-SAM \cite{li2023semanticsam} to pinpoint potential areas of distortion; 2) Provision of a textual quality evaluation message during the annotation process to serve as a reference. 
Our dataset is constructed on top of Q-Instruct \cite{wu2023qinstruct}, which provides detailed textual explanations regarding the image quality. 
Consequently, we have compiled a visual quality grounding dataset containing 50K human-annotated triplet samples \textit{(image, quality text, distortion segmentation)}. Recognizing the time-consuming and costly nature of human annotation,  we additionally automate part of the dataset creation using GPT4V \cite{gpt4v} because of its superior performance in overall quality evaluation \cite{wu2023qbench}. By employing the set-of-mark strategy \cite{yang2023setofmark}, we manage to collect an additional 50K samples for our dataset. These automatically labeled data can be easily generated, significantly broadening the diversity of our dataset. These two parts form our final dataset, \textbf{\dataset}, the first visual quality grounding dataset for fine-scale IQA.

The \dataset dataset enables the training of a quality grounding model for IQA. Rather than constructing a traditional visual grounding model that relies on separate embeddings for text and image as inputs, our aim is to develop a more capable and flexible multi-modality model that incorporates both text and images as inputs and outputs, akin to recent LMMs \cite{llava,mplugowl,lai2023lisa,ren2023pixellm}. 
Different from these existing methods, which primarily address high-level concepts, the visual quality grounding task places a greater emphasis on low-level and mid-level details. 
Consequently, we introduce a multi-scale feature abstractor (MSFA) to get quality-aware visual embeddings before merging them with text embeddings into pretrained large language model (LLM), thereby augmenting LMM's capacity for low-level perception. Furthermore, we train our model using a diverse dataset comprising high-level multi-modality data, the quality reasoning dataset \cite{wu2023qinstruct}, and the newly proposed \dataset. 
These varied datasets enable our model to undertake complex tasks, such as answering text-based questions about image content and quality, as well as conducting distortion segmentation. By integrating these features, our approach smoothly combines fine-scale and overall quality perception capabilities within the interactive analysis of visual contents, setting a new benchmark for future explorations in the field.

Our contributions can be summarized as follows:

\begin{itemize}
    \item To the best of our knowledge, we are the first to present framework aimed at fine-scale visual quality grounding, using the strengths of LMMs for detailed visual quality analysis.
    \item We construct the \dataset dataset, the first-of-its-kind dataset comprising 100K samples designed to support deep investigations into visual quality, encompassing both human-labeled and LMM-generated annotations.
    \item We introduce multi-scale visual feature abstractor for LMM-based visual quality grounding. The model is capable of performing image quality assessment and distortion segmentation with textual prompts, thus facilitating a fine-scale understanding of quality and interactive engagement with visual content.
    \item Our work establishes a new benchmark for future research in IQA, paving the way for more sophisticated and fine-grained analyses of image quality.
\end{itemize}

\section{Related Works}

\subsection{Image Quality Assessment}

\subsubsection{Previous Methods}

Current methods in IQA can be broadly divided into Full-Reference (FR) and No-Reference (NR) techniques. FR methods assess the discrepancy between a reference image and its distorted counterpart. The widely recognized Peak Signal-to-Noise Ratio (PSNR) evaluates this difference on a pixel-wise basis, whereas the Structural Similarity Index (SSIM) \cite{ssim,ssim2004} enhances this evaluation by incorporating structural similarity features, thereby inspiring several subsequent studies \cite{sheikh2006vif,cwssim2009,larson2010madcsiq,zhang2011fsim,xue2013gmsd,zhang2014vsi,laparra2016nlpd}. Learning-based approaches \cite{kim2017deepqa,bosse2017wadiqam,zhang2018lpips,Prashnani_2018_PieAPP,cheon2021iqt,topiq} have come to dominate FR IQA with significantly better performance, providing more accurate and reliable assessments of image quality. However, the necessity for a reference image limits their applications.

The development of the more challenging NR-IQA has followed a similar trajectory to that of FR-IQA. 
Traditional methods, exemplified by NIQE \cite{niqe}, rely on natural scene statistics \cite{moorthy2011nssdiivine,2012brisque,zhang2015ilniqe,ma2017nrqm,blau2018pi,tpqi}. 
In contrast, recent advancements \cite{hyperiqa,dbcnn,musiq,sun2023stairiqa,fastvqa,xu2023local,wang2023deep} in deep learning enable methods to directly learn to estimate MOS in an end-to-end fashion. Wang \etal \cite{wang2021active,wang2021troubleshooting} introduce gMAD examples to improve IQA performance. The efficacy of these deep learning models is closely related to the datasets they are trained on, resulting in capabilities that are less interpretable. For instance, a model trained on an aesthetic assessment dataset may excel at evaluating aesthetic quality \cite{nima,hou2022kd}, yet recognizing this specialty from its output scores is not straightforward. 
The emergence of multi-modality models, notably CLIP \cite{clip}, has inspired recent initiatives \cite{clipiqa,vila,liqe,clipiaa} to integrate the descriptive power of textual information with IQA, proving beneficial. 
Consequently, the latest works \cite{wu2023qinstruct,wu2023qalign,huang2024visualcritic,wu2024comprehensive,wu2024openended,depictqa} employ LMMs in IQA, significantly enhancing both performance and interpretability, and leading to a new era in IQA research. 
Despite these significant advancements, current IQA methods are limited to providing either a global score or a textual evaluation and lack the capability to evaluate image quality within the context of local distortions. Our work aims to bridge this gap.

\begin{figure*}[!t]
    \centering
    \includegraphics[width=\linewidth]{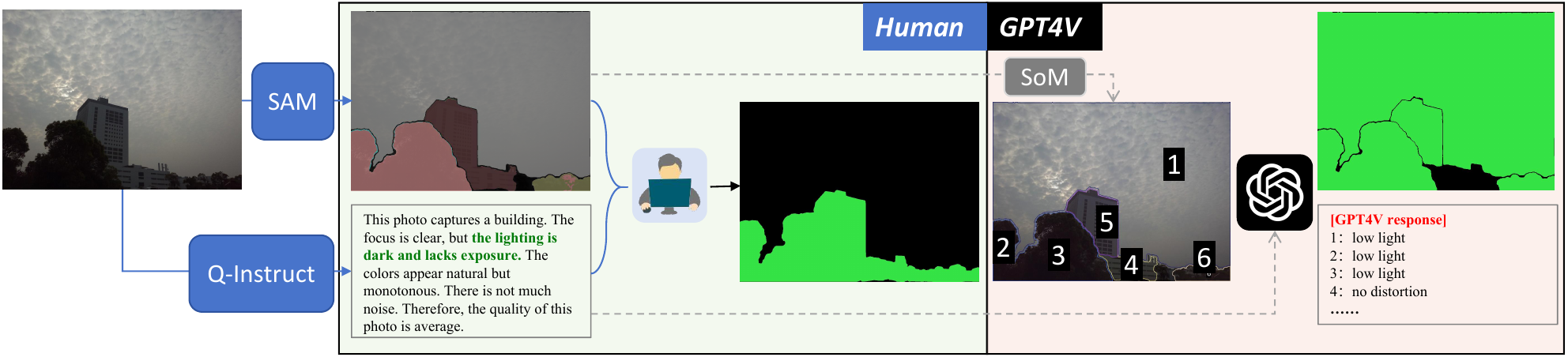}
    \caption{The data annotation pipeline incorporates both human expertise and GPT-4V capabilities. Firstly, the input image undergoes pre-segmentation using SAM \cite{li2023semanticsam}. In the human annotation phase, subjectives need to identify and categorize types of distortions, with quality description texts from humans as reference. The subjective is free to adjust borders generated by SAM. In the GPT4V annotation phase, the reference for quality is generated by the Q-Instruct model. Then, each region is marked with a number, which is then coupled with the quality text and forwarded to the GPT4V model. Finally, the model outputs the types of distortions present in each specified region.
    }
    \label{fig:data_anno}
\end{figure*}

\begin{table}[t]
    \centering
    \renewcommand\tabcolsep{6pt}
    \renewcommand\arraystretch{1.3}
    \caption{Comparison of existing public IQA datasets and the proposed QGround-100K.}
    \label{tab:intro}
    \begin{tabular}{c|c|cccc}
    \hline
     Type & Dataset & MOS & Text & Seg  \\ \hline
    FR & \makecell[c]{Traditional datasets \\ \cite{larson2010madcsiq} \cite{sheikh2006liveiqa} \cite{ponomarenko2009tid2008} \cite{kadid10k} \cite{zhang2018lpips}} & \cmark & \xmark & \xmark \\ \hline
     \multirow{3}{*}{NR} & \makecell[c]{Traditional datasets \\ \cite{livechallenge} \cite{koniq10k} \cite{fang2020spaq} \cite{ava} \cite{flivepaq2piq} } & \cmark & \xmark & \xmark  \\ \cdashline{2-5}
     & Q-Instruct \cite{wu2023qinstruct} & \cmark & \cmark & \xmark \\ 
     & \textbf{\dataset} & \cmark & \cmark & \cmark 
    \\ \hline
    \end{tabular}
\end{table}

\subsubsection{Existing Datasets} 

There are numerous datasets that have been pivotal in the development of both FR and NR IQA algorithms, as summarized in \cref{tab:intro}. 
The FR datasets typically include images with synthetic distortions like Gaussian blur and white noise, where subjects compare two images and assign a quality score, which is a process that can introduce score ambiguities.
To address this, BAPPS \cite{zhang2018lpips} introduces a two-alternative forced choice to reduce score uncertainty. 
Traditional NR datasets typically require subjects to provide a simple score. SPAQ \cite{fang2020spaq} further requires quality ratings specific to various distortions and contents. While these datasets are invaluable for training and benchmarking IQA models, the reliance on simple quality scores limits their interpretability. 
Therefore, Q-Instruct \cite{wu2023qinstruct} introduces textual quality descriptions, significantly enhancing the interpretability of IQA datasets. 
Nevertheless, these datasets mainly focus on global quality assessments, paying less attention to local distortion identification and detailed quality analysis. 
This limitation narrows their utility in applications demanding precise local distortion analysis, such as in image enhancement and editing tasks. 
The proposed \dataset dataset seeks to bridge this gap with comprehensive annotations including MOS, textual evaluations, and segmentation masks, establishing a more versatile tool for advanced IQA applications.

\subsection{Visual Grounding with LMM}

Visual quality grounding has long been an important task in computer vision, serving as a bridge between visual data and textual descriptions. 
Prior visual grounding, also known as referred expression comprehension, is mostly like a text conditioned localization task, see \cite{qiao2020referring} for a comprehensive survey. 
The evolution of LMMs has significantly influenced recent developments. Innovations such as Kosmos-2 \cite{kosmos2}, Shikra \cite{shikra}, GPT4RoI \cite{zhang2023gpt4roi}, VisionLLM \cite{wang2024visionllm} \etc, have successfully merged generative LMMs with localization tasks, facilitating human-model interactions at the region level. 
Recent advancements, notably LISA \cite{lai2023lisa}, GLaMM \cite{hanoona2023GLaMM}, and PixelLM \cite{ren2023pixellm}, have significantly improved upon existing methods by introducing pixel-level segmentation. However, the application of LMMs in the specific context of image quality assessment and visual grounding remains relatively unexplored. Our work takes a pioneering step forward in advancing fine-grained quality perception, marking a notable contribution to this evolving landscape.

\section{The \dataset Dataset}

In this section, we provide details about the process of constructing the \dataset dataset, which lays the foundation for enabling visual quality grounding. We discuss the sources of our data in \cref{sec:data_collect}, and outline the annotation pipeline involving both human annotators and GPT4V in \cref{sec:data_ann}. Additionally, we provide an analysis and statistics of labels obtained from human annotators and GPT4V in \cref{sec:data_stats}, offering insight into the dataset's composition and the reliability of its annotations.

\subsection{Data Collection} \label{sec:data_collect}

To develop a model capable of visual quality grounding, a dataset comprising triplet samples is essential: an input image, associated quality descriptive text, and ground truth distortion segmentation masks. 
Since the Q-Instruct \cite{wu2023qinstruct} dataset provides comprehensive text descriptions for images from diverse resources, we choose to build \dataset upon it, as summarized in \cref{tab:data}. 
We exclude 1K synthetic distorted images from COCO due to their focus on global distortions. 
As outlined in \cref{tab:data}, for images within the Q-Pathway that already have human-labeled texts, we complement them with human labeled segmentation masks. 
To enrich the diversity of images, we include the rest images from IQA datasets for GPT4V labeling. Recognizing the rising popularity of AI-generated images, we also add 5.5K images from \cite{agiqa3k,imagereward}. The accompanying quality text is generated using latest Co-Instruct model\footnote{\url{https://huggingface.co/q-future/co-instruct}}, chosen for its performance comparable to that of GPT4V.    

\subsection{Data Annotation} \label{sec:data_ann}

To streamline the annotation process, we have chosen five prevalent types of distortions for mask annotation: \texttt{blur}, \texttt{overexposure}, \texttt{noise}, \texttt{jitter} and \texttt{low light}. These categories were selected for their frequency and significance in impacting visual quality across a wide range of images according to the report in Q-Instruct \cite{wu2023qinstruct}. 
\Cref{fig:data_anno} showcases the comprehensive data annotation pipeline, which incorporates both human and GPT4V annotation stages. Below, we provide detailed explanations for each phase within the pipeline, ensuring clarity and insight into our systematic approach for annotating the \dataset dataset. 

\begin{table}[t]
    \centering
    \caption{The image sources and statistics of \dataset.} \label{tab:data}
    \renewcommand\arraystretch{1.10}
    \renewcommand\tabcolsep{2pt}
    \begin{tabular}{c|ccc}
    \hline
     Image Sources &  Original & \makecell{Human labeled \\ (Q-Pathway)} & GPT4V-labeled \\ \hline
     KonIQ-10K~\citep{koniq10k} & 10,373 & 5,182 & 5,168 \\
     SPAQ~\citep{fang2020spaq} & 11,125 & 10,797 & --- \\
     LIVE-FB~\citep{flivepaq2piq} & 39,810 & 800 & 38,946 \\
     LIVE-itw~\citep{livechallenge} & 1,169 & 200 & 969 \\ 
     AGIQA-3K~\citep{agiqa3k} & 2,982 & 400 & 2,568 \\
     ImageRewardDB~\citep{imagereward} & 50,000 & 584 & 2,947 \\ \hline
     \# Image & -- & 17,963 & 50,599 \\  
     \# Annotation & -- & 52,924 & 50,599 \\  
    \hline
    \end{tabular}
    \label{tab:my_label}
\end{table}

\subsubsection{Human Annotation} 
In the human annotation phase, 15 trained annotators with solid educational backgrounds are presented with (image, quality description) pairs. Their task is to segment out the distorted regions within the images and categorize the types of distortions present.
To minimize ambiguity in the annotations, annotators are instructed to consult the provided quality descriptions throughout the annotation process. This step is crucial to ensure that the regions of interest related to quality assessment are accurately highlighted. 
Additionally, a pre-segmentation step utilizing SAM \cite{li2023semanticsam} is implemented to improve uniformity in the segmentation boundaries associated with specific distortions. Despite this automated assistance, annotators retain the judgement to manually adjust the boundaries. This flexibility acknowledges that SAM's segmentation may prioritize object areas over actual distortion locations, thereby allowing for more precise identification of quality-related distortions. 

\subsubsection{GPT4V Annotation}
Following the human annotation phase, the GPT-4V annotation employs the Set-of-Mark (SoM) technique to facilitate mask annotation. 
As depicted in \cref{fig:data_anno}, the images are initially segmented using SAM and labeled with numbers. Subsequently, GPT4V is provided with the same (image, quality description) pairs that were utilized in the human annotation process. 
Leveraging its profound comprehension of both visual and textual content, the model identifies and labels regions of distortion.
This methodology allows GPT4V to autonomously generate segmentation masks for distorted regions within an image, informed by the provided quality descriptions. 
This automated process not only speeds up the annotation effort but also provides a scalable way to enrich the dataset with diverse interpretations of image quality, bridging the gap between human efforts and AI efficiency.

\subsection{Analysis of \dataset} \label{sec:data_stats}

\begin{figure}[t]
    \centering
    \begin{subfigure}{\linewidth}
        \includegraphics[width=\linewidth]{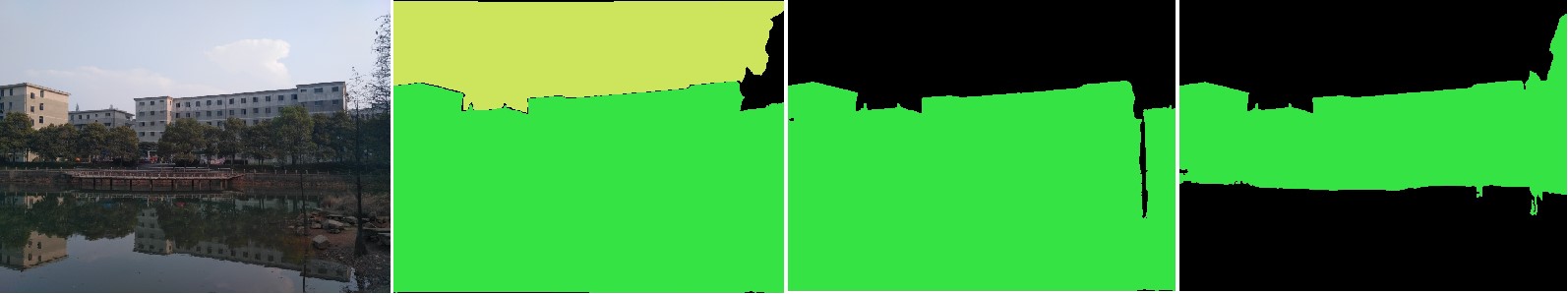}
        \caption{Example of agreement among three annotations.} \label{fig:data_agree_1}
    \end{subfigure}
    \begin{subfigure}{\linewidth}
        \includegraphics[width=\linewidth]{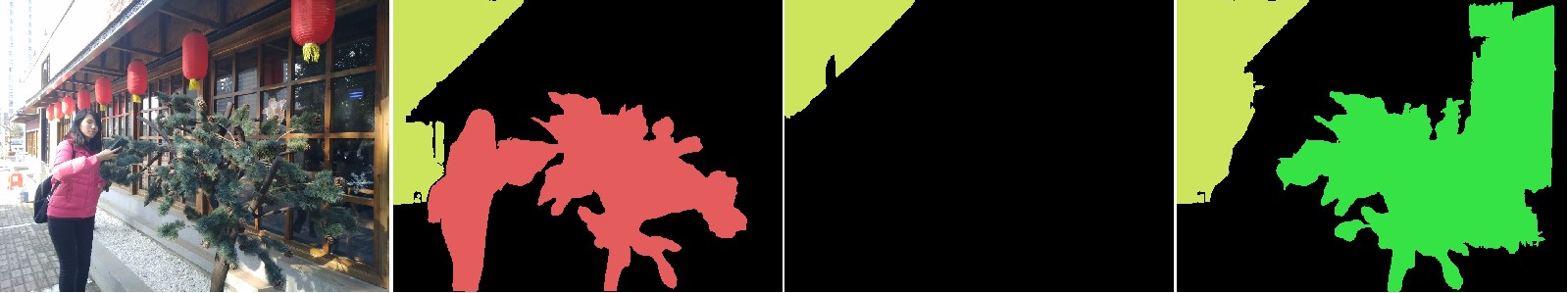}
        \caption{Example of disagreement among three annotations.} \label{fig:data_agree_2}
    \end{subfigure}
    \begin{subfigure}{\linewidth}
        \renewcommand\tabcolsep{1pt}
        \renewcommand\arraystretch{1.20}
        \resizebox{\linewidth}{!}{
            \begin{tabular}{c|c|c|c|c|c|c}
            \hline
             Dataset & KonIQ-10K & SPAQ & LIVE-FB & LIVE-itw & AGIQA-3K & ImageReward \\ \hline 
             Recall & 0.902 & 0.864 & 0.931 & 0.966 & 0.976 & 0.980 \\ \hline
            \end{tabular}
        }
        \caption{Pairwise recall between different annotators on different datasets.} \label{fig:data_agree_3}
    \end{subfigure}
    \caption{Analysis of annotation agreement between different human subjectives.}
    \label{fig:data_agree}
\end{figure}

\begin{figure}[t]
    \centering
    \includegraphics[width=\linewidth]{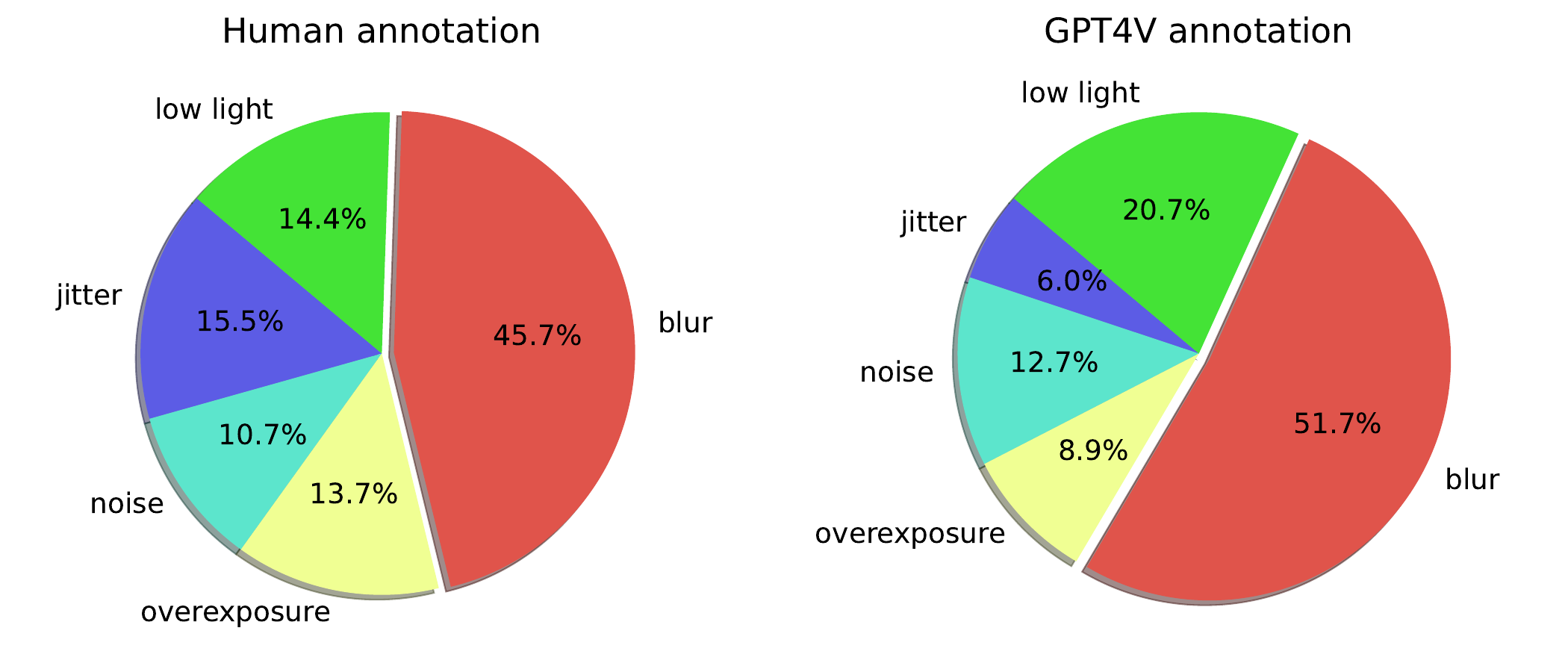}
    \caption{Statistics of human and GPT4V parts separately.}
    \label{fig:data_stats}
\end{figure}

\begin{figure*}[!t]
    \centering
    \includegraphics[width=\linewidth]{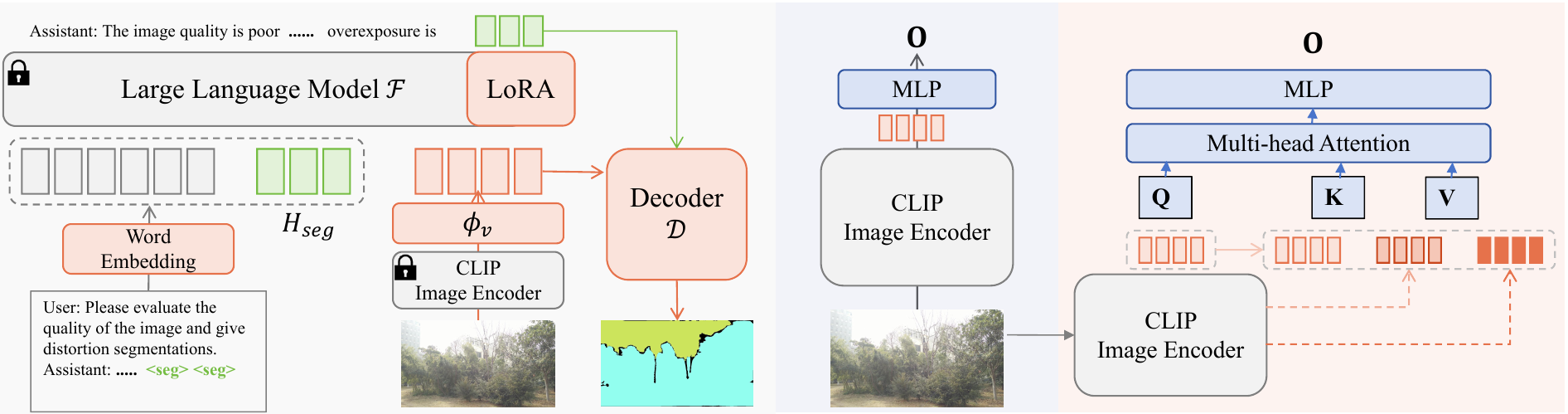}
    \makebox[.5\linewidth]{(a) Overall framework}
    \makebox[.15\linewidth]{(b) $\phi_v$ in \cite{lai2023lisa,ren2023pixellm}}
    \makebox[.34\linewidth]{(c) Proposed Multi-scale Feature Abstractor}
    \caption{The pipeline of our method. (a) The framework follows previous methods \cite{lai2023lisa,ren2023pixellm} and is designed to accept inputs of images and texts, subsequently producing textual outputs and segmentation results. (b)(c): comparison of multi-modal projection block between previous works and our proposed multi-scale feature abstractor.}
    \label{fig:method_arch}
    \vspace{-1em}
\end{figure*}

As summarized in \cref{tab:data}, the \dataset dataset comprises rich images and annotations for visual quality grounding from both human and GPT4V. 
Here, we delve into the statistics between human and GPT4V annotations.

Given that the range and types of distortions can be subjective and may vary among different annotators, assessing the reliability of human annotations is crucial. To this end, we examine the agreement between various human annotations within the Q-Pathway dataset, where each image is associated with at least three distinct quality text annotations. Different annotators label the same image but with different accompanying texts. As depicted in \cref{fig:data_agree}, we consider the results acceptable when one mask is a subset of another (\cref{fig:data_agree_1}), and unacceptable when the same region is labeled with different types of distortions (\cref{fig:data_agree_2}).
To quantitatively evaluate the agreement score of human annotations, we employ the recall of the pairwise intersection area over the smaller masks as follows:
\begin{equation}
    \textup{Recall} = \frac{1}{N} \sum_i^N \Bigl [ \frac{1}{M_i} \sum_j^{M_i} \frac{A_j \cap B_j}{\min(A_j, B_j)}\Bigr],
\end{equation}
where $M_i = {2 \choose m_i}$, $N$ is the number of images, $m_i$ is the number of masks for image $i$ and $A, B$ are the selected pairs. The findings, illustrated in \cref{fig:data_agree_3}, reveal that the agreement scores across different datasets are notably high, underscoring the reliability of our human annotation process.
Regarding GPT4V annotations, it was observed that GPT4V consistently yields similar results across multiple runs when provided with appropriate prompts (detailed further in the supplementary materials). 
This analysis confirms the robustness and dependability of both human and GPT4V annotations within our dataset, laying a strong foundation for accurate visual quality grounding.

Besides, we provide an analysis of the distribution of distortion types found within both human and GPT4V annotations, as detailed in \cref{fig:data_stats}. Generally, the frequency of distortions observed roughly follows the order: \texttt{blur} > \texttt{low light} > \texttt{overexposure} $\approx$ \texttt{noise}. A notable deviation in this pattern is the higher incidence of \texttt{jitter} in human annotations compared to those by GPT4V. 
This difference likely comes from the Q-Pathway's substantial inclusion of images from SPAQ \cite{fang2020spaq}, a dataset composed of smartphone-captured images, which are prone to \texttt{jitter} due to hand movement. Conversely, the segment annotated by GPT4V primarily consists of web-crawled images, where \texttt{jitter} is less common.
The similarity in the distribution of distortion annotations between human annotators and GPT4V highlights the GPT4V's effectiveness as a data generator, thereby validating its use in supplementing and expanding the dataset. 
Overall, the combined efforts of human and GPT4V annotations significantly enhance the diversity and utility of the dataset, providing a promising way to scale up datasets for visual quality grounding.

\section{Methodology}

\begin{figure*}[!t]
    \centering
    \includegraphics[width=\linewidth]{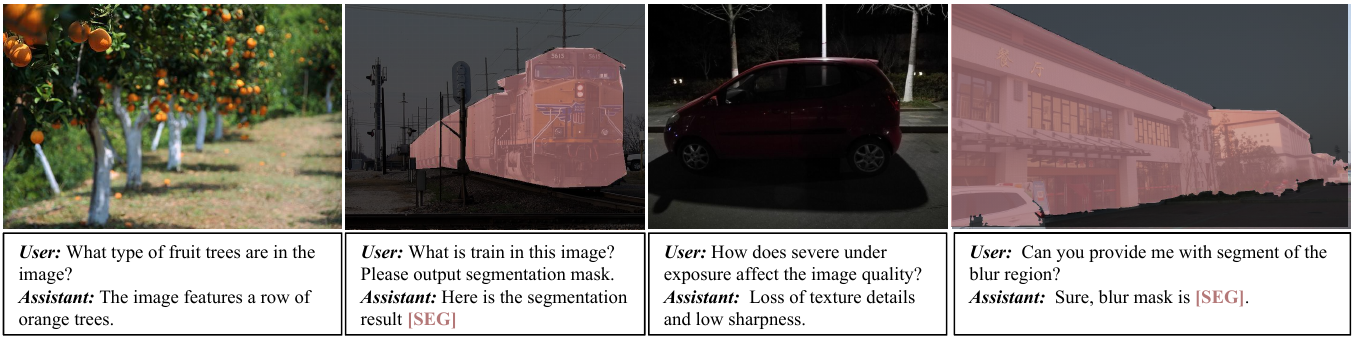}
    \makebox[0.24\linewidth]{Visual question answering.}
    \makebox[0.22\linewidth]{Semantic segmentation.}
    \makebox[0.22\linewidth]{Visual quality reasoning.}
    \makebox[0.28\linewidth]{Visual quality grounding.}
    \vspace{-1em}
    \caption{Example from various data sources for multi-task training of Q-Ground model.}
    \label{fig:data_multitask}
    \vspace{-1em}
\end{figure*}

Our objective is to develop a model capable of dialogues with users concerning the content and quality of images, while also executing distortion region segmentation through text queries. 
Our framework follows the simple and efficient pipeline of PixelLM \cite{ren2023pixellm} (\cref{sec:framework}), and we improve the image to text projection block with multi-scale features to enhance quality-aware perception of the model (\cref{sec:msfa}). 
Then, we train the model with multi-task datasets to enhance its capabilities (\cref{sec:multitask}).  

\subsection{The Overall Framework} \label{sec:framework}

As illustrated in \cref{fig:method_arch}(a), the system processes both image inputs, denoted as $x_{img}$, and textual inputs, $x_{txt}$, to produce corresponding textual responses, $y_{txt}$, and segmentation masks, $y_m$. The inputs $I$ and $x_{txt}$ are firstly transformed into token embeddings, which are subsequently processed by a pre-trained large language model (LLM), such as LLaMA \cite{llama}, to generate output tokens in an auto-regressive manner. These tokens are then decoded to form $y_{txt}$. To facilitate the generation of segmentation outputs, we draw inspiration from previous works \cite{lai2023lisa,ren2023pixellm} and introduce learnable segmentation tokens, represented as $H_{seg} = \{h^i \in \mathbb{R}^d\}|_{i=1}^N$, where $N$ represents the number of segmentation tokens and $d$ indicates the dimension of features. The segmentation masks, $y_m$, are generated using a decoder that takes the embeddings of $x_{img}$ as image input and $C_{seg}$ as condition inputs. This process involves the use of the pre-trained LLM, denoted as $\mathcal{F}$, and the CLIP image encoder, represented as $\mathcal{V}$. The overall pipeline of our framework is thus formulated as follows:
\begin{align}
    \{y_{txt}, H_{seg}\} &= \mathcal{F}(\phi_v(\mathcal{V}(x_{img})), x_{txt}, H_{seg}), \\
    y_{seg} & = \mathcal{D}(\mathcal{V}(x_{img}), H_{seg}),  
\end{align}
where $\mathcal{D}$ is a mask decoder same as \cite{ren2023pixellm}, and $\phi_v$ is the projector from visual features to LLM embedding space. 

As shown in \cref{fig:method_arch}(a), prior studies \cite{lai2023lisa,ren2023pixellm} typically select straightforward Multilayer Perceptron (MLP) as $\phi_v$ and only use the final features from $\mathcal{V}(x_{img}) \in \mathbb{R}^{(h\times w) \times d_v}$, focusing mainly on high-level representations. Nonetheless, in our task centered on visual quality grounding, multi-scale features are critical for learning quality-associated perceptions, as evidenced by previous research \cite{topiq}. Therefore, we introduce an innovative approach for $\phi_v$ that incorporates multi-scale features, as elaborated below.

\subsection{Multi-scale Feature Abstractor} \label{sec:msfa}

The architecture of our proposed Multi-Scale Feature Abstractor (MSFA) is depicted in \cref{fig:method_arch}(c). 
Modern vision encoders mainly employ a vision transformer structure, exemplified by ViT/14
For an image with size $H\times W$, the feature dimensions from different layers remain the same as $\frac{H}{14}\times \frac{W}{14}$. 
A straightforward solution is to directly put multi-scale features into LLM, which will significantly increases computational cost due to the exponential rise in attention calculation as token length extends. For example, when $H=W=448$ and 3 scales are used, the visual token length alone would be as long as $1024\times3$. 
On the other hand, such extensive visual tokens may not be essential owing to the redundancy in visual features. Recent study \cite{xiao2023semantic} shows that $256$ tokens might be enough for integrating image features with LLM.
Therefore, we present a multi-scale feature abstractor that employs a fixed-length query to distill useful information from multi-scale features. 
Given a set of multi-scale features $\mathbf{F}=\{f_i \in \mathbb{R}^{P\times d_v}\}$, where $f_i$ is the $i$-th layer feature from $\mathcal{V}(x_{img})$, the proposed MSFA can be calculated as
\begin{align}
    \mathbf{V} &= \textup{MHA}(\mathbf{Q}, \mathbf{F}, \mathbf{F}), \\
    \mathbf{O} &= \sigma(\mathbf{V} W_1)W_2, 
\end{align}
where the $\textup{MHA}$ denotes multi-head attention, $\sigma$ is the activation function, $W_1, W_2$ are parameters of linear layers, and the query feature $\mathbf{Q} \in \mathbb{R}^{256\times d_v}$. To simplify training, we use a pooled feature from the last layer of $\mathcal{V}(x_{img})$ as $\mathbf{Q}$, and $\mathbf{F}$ includes the last layer features in addition to features from several shallower layers.

\subsection{Multi-task Training} \label{sec:multitask}

To acquire a powerful LMM model capable of integrating visual quality grounding into interactive dialogues with users, we use a variety of publicly available data sources, as illustrated in \cref{fig:data_multitask}. Our training dataset consists of four parts, detailed as below:

\begin{itemize}[leftmargin=*]
    \item \textit{Visual question answering dataset.} This dataset enhances the model's understanding of visual content via question and answer pairs about the input image. We employ the LLaVA-Instruct-150K dataset \cite{llava} directly.
    \item \textit{Semantic segmentation dataset.} A collection used to preserve the semantic segmentation ability of the model, avoiding model overfitting to the distortion task. We include many different datasets for this part, \ie, ADE20K \cite{zhou2017scene}, COCO images \cite{lin2014coco}, COCO-stuff \cite{cocostuff}, as well as reasoning segmentation datasets from \cite{lai2023lisa,ren2023pixellm}. 
    \item \textit{Visual quality reasoning dataset.} The Q-Instruct dataset \cite{wu2023qinstruct} helps the model answer questions about visual quality.
    \item \textit{The proposed \dataset dataset.} Our uniquely compiled dataset, specifically designed to train the model on visual quality grounding in conversational contexts, enriching its ability to discuss about image content and quality.
\end{itemize}
Such diverse datasets contribute to a comprehensive understanding of visual content, quality assessment, and interactive communication, making our model promising for real-world applications.

\noindent \textbf{Training objectives.} The model produces both textual outputs and segmentation masks, employing auto-regression to train the text generation component and supervised learning for the segmentation mask. In line with prior research, we apply two distinct loss functions for each output: cross-entropy loss for text generation and a hybrid of binary cross-entropy and DICE loss for mask creation. The overall loss function is represented as follows:
\begin{equation}
    \mathcal{L} = \lambda_{txt} \mathcal{L}_{ce}(y_{txt}, \hat{y}_{txt}) + \lambda_{seg} \mathcal{L}_{seg}(y_{seg}, \hat{y}_{seg}),
\end{equation}
where $\hat{y}_{txt}$ is the shifted texts, $\hat{y}_{seg}$ is the ground truth mask, and $\lambda$ are loss weights. More details are given in supplementary material.

\section{Experiments} 

\subsection{Implementation Details}

\begin{figure*}[t]
    \centering
    \newcommand{\imgwidth}{0.164\linewidth}
    \includegraphics[width=\linewidth]{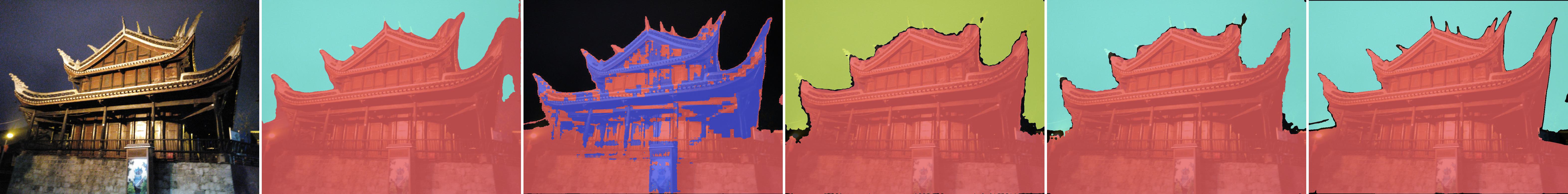}
    \includegraphics[width=\linewidth]{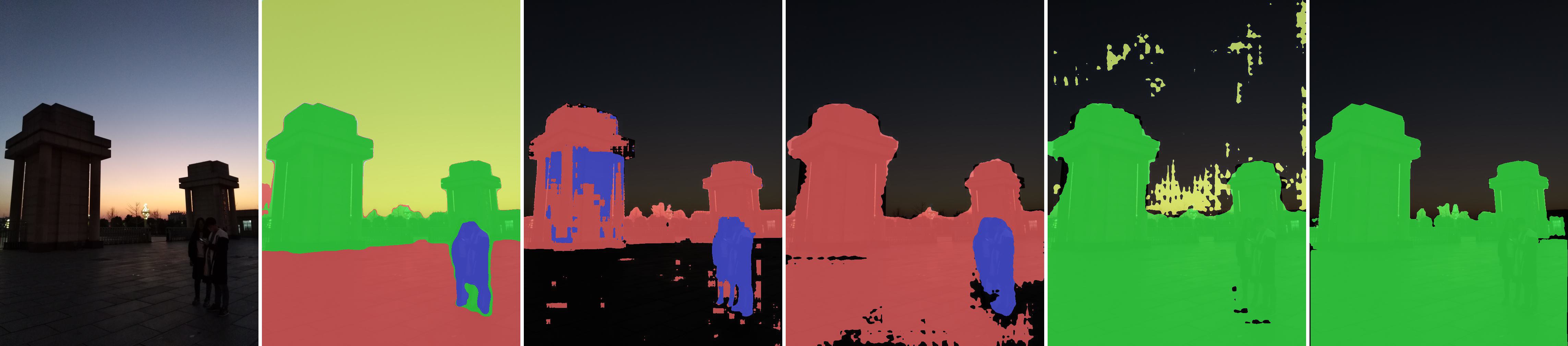}
    
    \makebox[\imgwidth]{Input}
    \makebox[\imgwidth]{Mask2Former \cite{cheng2021mask2former}}
    \makebox[\imgwidth]{LISA \cite{lai2023lisa}}
    \makebox[\imgwidth]{PixelLM \cite{ren2023pixellm}}
    \makebox[\imgwidth]{Ours}
    \makebox[\imgwidth]{Ground Truth}
    
    \vspace{-1em}
    \caption{Visual comparison of segmentation results for distortions: \colorbox{jitter}{\strut jitter}, \colorbox{noise}{\strut noise}, \colorbox{overexposure}{\strut overexposure}, \colorbox{blur}{\strut blur} and \colorbox{lowlight}{\strut low light}.}
    \label{fig:vis_results}
    \vspace{-1em}
\end{figure*}

\subsubsection{Training Details} 
Our model is finetuned from the pretrained LLaVA-7B model \cite{llava}, with CLIP-ViT-L/14-336 for visual encoding. To enhance detail capture, we follow \cite{ren2023pixellm} and resize the input image to $448\times448$. 
The trainable modules include the word embedding, LoRA parameters for LLM, visual projector $\phi_v$ and the mask decoder $\mathcal{D}$. 
We employ the AdamW \cite{adamw} optimizer, setting the learning rate at 0.0003, and utilize the WarmupDecayLR scheduler, which begins with 100 warmup iterations.
The batch size is set to 2 per device with 10 steps of gradient accumulation. The model is firstly pretrained with semantic segmentation datasets to obtain common semantic abilities and then finetuned with \dataset dataset for visual quality grounding. The total training process requires approximately 2 days on 4 NVIDIA 4090 GPUs.

\subsubsection{Benchmark Dataset and Evaluation Metrics}
As a new task, we establish a new benchmark for evaluating visual quality grounding. As detailed in \cref{tab:data}, the proposed \dataset comprises $17,963$ unique images, each annotated with human-labeled masks. We randomly split $1,000$ as the test set. Each image is accompanied by a minimum of three distinct quality descriptions and may be associated with up to three different ground truth masks. 

For quantitative evaluation, we rely on the widely recognized metrics for segmentation task, \ie, the mean Intersection over Union (mIoU) and mean classification accuracy (mAcc).

\subsection{Benchmark Performance}

\subsubsection{Selected Methods and Evaluation Protocal}

Since visual quality grounding is a new task for image quality assessment, there is no existing works to compare directly as far as we know. We therefore select two kinds of methods that are closely related:
\begin{itemize}[leftmargin=*]
    \item \textbf{Semantic segmentation.} We select two exemplary segmentation techniques, \ie, SegFormer \cite{xie2021segformer} and Mask2Former \cite{cheng2021mask2former}, along with a recent open-vocabulary model, SAN \cite{xu2023san}, as representative methods for our analysis. Given that these models do not process textual inputs and are capable of producing only a single outcome per input image, we calculate their average performance since there are multiple ground truth masks for one input image.
    \item \textbf{LMM based reasoning segmentation.} This area of study is relatively new and closely aligns with our work. We choose two of the most recent contributions, LISA \cite{lai2023lisa} and PixelLM \cite{ren2023pixellm}, as methods for comparison.
\end{itemize}

Since methods based on LMMs accommodate flexible inputs and outputs, for a fair comparison, we evaluate each method using prompts like: \prompt{<quality text> Please segment out distorted regions in the image.} to obtain the corresponding mask for the identified distortions, where \prompt{<quality text>} is the global quality reasoning text. We use \textit{``smaller region first''} principle to merge various segmentation masks in the event of overlaps, because it prioritizes precision and diversity in segmentation, ensuring that more details are captured and evaluated. \emph{All these compared methods are re-trained or finetuned with \dataset dataset.}

\begin{table*}[t]
\caption{Quantitative comparison with segmentation methods and LMM-based methods on QGround-Test. The averages are weighted based on the number of images
with different distortions.} \label{tab:quant_results}
\renewcommand\tabcolsep{7pt}
\renewcommand\arraystretch{1.10}
\begin{tabular}{c|cc|cc|cc|cc|cc|cc}
\hline
\multirow{2}{*}{Method} & \multicolumn{2}{c|}{\cellcolor{jitter} jitter} & \multicolumn{2}{c|}{\cellcolor{noise} noise} & \multicolumn{2}{c|}{\cellcolor{overexposure} overexposure} & \multicolumn{2}{c|}{\cellcolor{blur} blur} & \multicolumn{2}{c|}{\cellcolor{lowlight} low light} & \multicolumn{2}{c}{Average} \\ \cline{2-13}
 & mIoU & mAcc & mIoU & mAcc & mIoU & mAcc & mIoU & mAcc & mIoU & mAcc & mIoU & mAcc \\ \hline
 SegFormer \cite{xie2021segformer} & 0.327 & 0.625 & 0.136 & 0.249 & 0.264 & 0.389 & 0.515 & 0.842 & 0.274 & 0.524 & 0.373 & 0.636 \\
 Mask2Former \cite{cheng2021mask2former} & 0.401 & 0.625 & 0.089 & 0.113 & 0.223 & 0.424 & 0.566 & 0.902 & 0.290 & 0.461 & \textbf{0.403} & \textbf{0.646} \\
 SAN \cite{xu2023san} & 0.119 & 0.239 & 0.011 & 0.018 & 0.143 & 0.454  & 0.387 & 0.584 & 0.162 & 0.223 & 0.228 & 0.401 \\ \hdashline
 LISA \cite{lai2023lisa} & 0.154 & 0.688 & 0.003 & 0.003 & 0.082 & 0.102 & 0.411 & 0.682 & 0.005 & 0.006 & 0.227 & 0.436 \\ 
 PixelLM \cite{ren2023pixellm} & 0.400 & 0.823 & 0.050 & 0.200 & 0.117 & 0.380 & 0.429 & 0.632  & 0.131 & 0.185 & 0.252 & 0.519 \\ 
 Ours &  0.434 & 0.720 & 0.051 & 0.176 & 0.125 & 0.459 & 0.460 & 0.648 & 0.219 & 0.337 & \textbf{0.271} & \textbf{0.539} \\ \hline
\end{tabular}
\end{table*}

\subsubsection{Results Comparison on QGround Benchmark}

According to the results shown in \cref{fig:vis_results} and \cref{tab:quant_results}, we can notice the difference in performance between semantic segmentation models and LMM-based approaches. The traditional semantics segmentation model, especially Mask2Former, generates masks with better details and cleaner boundaries and the quantitative performance is also better. The exception, SAN, is worse likely due to its optimization for high-level segmentation tasks and lack of suitable mask decoder for visual quality grounding. 
The superior performance of segmentation methods is probably because they are better at the simple five-class segmentation task.
Meanwhile, the LMM-based approaches face the dual challenge of identifying distortion types while concurrently generating segmentation results. Nevertheless, LLM-based methods demonstrate a significant advantage in versatility and capability over traditional segmentation techniques, offering additional abilities such as answering questions about image quality and content. 

In LMM based approaches, PixelLM and ours outperform LISA in mask classification.
This improvement is attributed to the benefit of optimizing multiple segmentation tokens, which enhances classification accuracy. 
On the other hand, the utilization of multi-scale features in visual projection further improves the quality concept understanding of LMM, leading to our superior performance compared with PixelLM. Examples in \cref{fig:vis_results} shows that improved local representation enables our proposed MSFA to better distinguish between noise and overexposure, as well as blur and low light distortions.

\subsection{Analysis and Ablation Study}

\begin{table}[t]
    \centering
    \caption{Ablation study of datasets used in training.}
    \resizebox{\linewidth}{!}{
        \begin{tabular}{c|ccc:cc|cc}
        \hline
        ID & Q \& A & Seg & Q-Inst & \makecell{QG-\\human} & \makecell{QG-\\GPT} & mIoU & mAcc \\ \hline
        I & \cmark & \cmark & & & & \color{lightgray} 0.042 & \color{lightgray} 0.113 \\ 
        II & \cmark & \cmark &  & \cmark & \cmark & 0.267 & 0.538 \\ 
        III &  & & \cmark & \cmark & \cmark & \textbf{0.275} & \textbf{0.546} \\ \hdashline
        IV & \cmark & \cmark & \cmark & \cmark & & 0.260 & 0.531 \\ 
        V & \cmark & \cmark & \cmark & \cmark & \cmark & 0.271 & 0.539 \\ \hline
        \end{tabular}
    }
    \label{tab:ablation_data}
\end{table}

\begin{table}[t]
    \centering
    \caption{Ablation studies. Left: scales used in Multi-Scale Feature Abstractor; right: quality text reference in prompt.}
    \label{tab:ablation_else}
    \renewcommand\tabcolsep{3pt}
    \begin{subtable}[t]{0.54\linewidth}
        \begin{tabular}[t]{r|cc}
        \hline
        Layers Used $\phi_v$ & mIoU & mAcc \\ \hline
        PixelLM (23) & 0.252 & 0.519 \\
        14, 23 &  0.269 & 0.538 \\
        7, 14, 23 & \textbf{0.271} & \textbf{0.539} \\
        \hline 
        \end{tabular}
    \end{subtable}
    \hfill
    \begin{subtable}[t]{0.44\linewidth}
        \begin{tabular}[t]{c|cc}
        \hline
        Txt Ref & mIoU & mAcc \\ \hline
        \xmark & 0.268 & 0.501 \\
        \cmark & 0.271 & 0.539 \\
        \hline 
        \end{tabular}
    \end{subtable}
    
\end{table}

\subsubsection{Dataset Fusion} We firstly examine the impact of mixed datasets training on the visual quality grounding results, as depicted in \cref{tab:ablation_data}. Experiment I employs no quality grounding data and serves as a foundational baseline.
From experiment II and III, it is observed that integrating tasks related to semantic segmentation shows little effect on the performance of quality grounding, while replacing them with Q-Instruct can produce marginally improved results. 
This suggests that these two tasks may be independent of one another, and their integration is feasible for developing a more capable model.
When comparing IV and V, it is evident that incorporating data labeled by GPT4V is beneficial to performance. We anticipate that incorporating GPT4V will prove even more beneficial in the context of more complex data annotation processes, a potential we intend to explore in future research.

\subsubsection{Multi-scale Feature Abstractor} 
\Cref{tab:ablation_else} demonstrates that incorporating mid-level features significantly enhances low-level perceptual capabilities, while the inclusion of shallower level features is also somewhat beneficial. Therefore, we empirically choose these three layers to achieve a good balance.

\subsubsection{Quality Text Reference in Prompt} 
\begin{figure}[t]
    \centering
    \dottedbox[black]{
    {\small \textbf{Quality text:} Overall, this image is not clear enough, the focus is not accurate enough, and the background content is \colorbox{blur}{too blurry}.}
    
    \includegraphics[width=\linewidth]{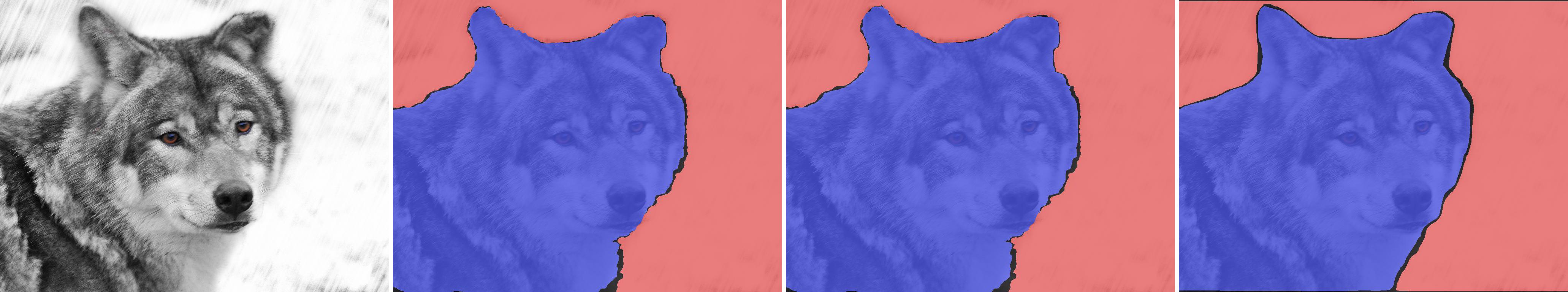}
    }
    \dottedbox[black]{
    {\small \textbf{Quality text:} The main subject of the image is a wolf, with overall poor sharpness, {\colorbox{overexposure}{average lighting}}, severe motion blur, unclear contours, and moderate texture clarity.}
    
    \includegraphics[width=\linewidth]{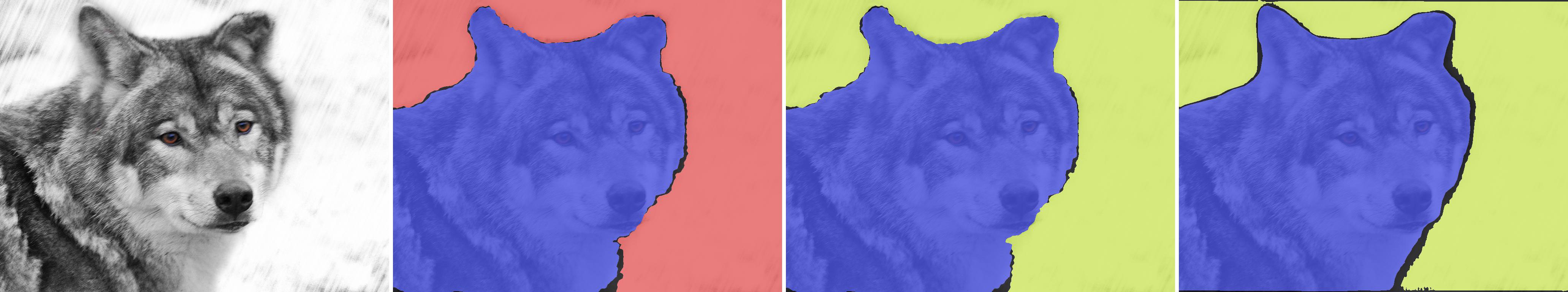}
    }
    \makebox[.24\linewidth]{\small Input Image}
    \makebox[.24\linewidth]{\small w/o quality text}
    \makebox[.24\linewidth]{\small w/ quality text}
    \makebox[.24\linewidth]{\small Ground Truth}
    
    \caption{Examples w/ and w/o quality text in prompt.}
    \label{fig:ablation_txt}
\end{figure}

\Cref{tab:ablation_else} also shows the significance of incorporating a global quality text reference.
The mAcc shows a considerable improvement compared to scenarios lacking a text reference. 
As illustrated in \cref{fig:ablation_txt}, the model can identify distortion types mentioned in the provided text and generates corresponding results, thereby facilitating more effective interaction with users.

\section{Conclusion}

In this study, we pioneer the integration of visual grounding into image quality assessment, enabling a more fine-grained perception of local quality. To accomplish this objective, we collected a comprehensive dataset comprising 100K annotated samples, namely, the \dataset. 
This dataset was carefully labeled, with half of the annotations provided by human participants and the remaining half by GPT4V, thereby enhancing both the diversity and efficiency of data labeling. 
With this \dataset, we introduced a LMM-based approach that seamlessly incorporates quality grounding within multi-modal tasks. 
Specifically, we developed a multi-scale feature abstractor (MSFA) designed to augment the LMM's capacity to recognize low-quality attributes.
Our research sets a new benchmark for the task of image quality assessment, broadening its potential applications across a wider range of fields.

\bibliographystyle{ACM-Reference-Format}
\bibliography{egbib}

\newpage
\appendix

\section{Data Collection Details}

In this section, we provide more details about data collection process including both human and GPT4V. 

\subsection{Human Annotation}

\subsubsection{Information of Participants}
We recruited 15 participants for this study, with a gender distribution of 10 females and 5 males. All participants are within the age range of 20 to 30 years and possess at least a college degree. The participants were selected to provide a diverse representation in terms of academic backgrounds, including disciplines such as computer science, psychology, and engineering.

\subsubsection{Preparation and Quality Control}
Participants underwent a comprehensive training session to familiarize them with the annotation guidelines and tools used in this study. The training included detailed explanations of the tasks and practice sessions to ensure clarity and consistency in the annotation process. 
The participants was trained on $1,000$ samples first, and the supervising teams checked the quality and improved the process.
To maintain high standards of annotation quality, we conducted periodic checks of the annotations during the annotation process.

\subsubsection{Annotation Pipeline} 
\begin{figure}[t]
    \centering
    \includegraphics[width=0.99\linewidth]{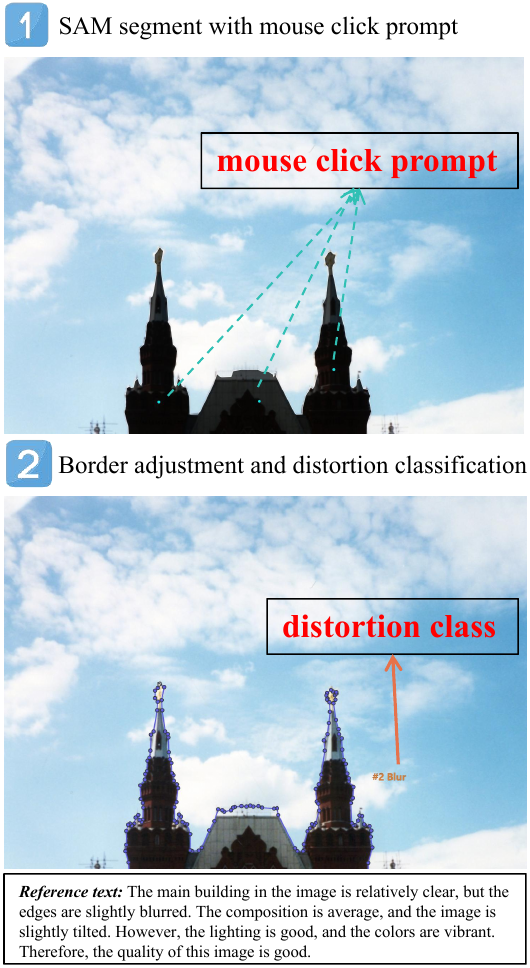}
    \caption{Example of human annotation steps with the help of SAM and reference quality text.}
    \label{fig:human_anno}
\end{figure}

\Cref{fig:human_anno} illustrates the annotation pipeline. To ensure simplicity and consistency in annotation, we have divided the process into two steps. 
Firstly, participants identify the distorted region by simply clicking on it. For instance, as shown in \cref{fig:human_anno}, the subject first reviews the reference text and determines whether the building is blurry. 
Upon identifying a blurry region, subjects merely click on that area; this action prompts the system to produce segmentation results using Semantic-SAM \cite{li2023semanticsam}. 
In the second step, subjects refine the annotations and assign distortion classification labels.

\subsubsection{Ethical Considerations}
All participants were informed about the goals of the research and the use of the annotated data. Consent was obtained from each participant, ensuring they understood their rights, including the right to withdraw from the study at any time without any consequences. Privacy and confidentiality of the participants were strictly maintained throughout the research process.

\subsection{GPT4V Annotation}

\begin{figure*}[!t]
\begin{tcolorbox}[title={Example of GPT4V Annotation with SoM (gpt-4-vision-preview)},colback=blue!5!white,colbacktitle=red!10!white,coltitle=black,fonttitle=\bfseries,colupper=blue]

\textit{\textbf{\#System:}} You are a helpful assistant to help me evaluate the quality of the image. The image is divided into several regions with number marks. You will be given an overall evaluation of the quality as reference. Please help to identify the distortions of each region within the following types [blur, jitter, overexposure, low light, noise, no distortion]. Please give the result in the following json format:
\begin{verbatim}
[{
    "[mark number]": "distortion type",
    "gpt4v iqa": "message",
}]
\end{verbatim}
Please note that the distortion type should be one of the five types mentioned above, and the message should be a brief evaluation of the quality of the region. Please strictly follow the format, otherwise the result will be invalid.

\tcblower
\begin{minipage}{0.49\linewidth}
\textit{\textbf{\#User:}} The overall quality reference is: \prompt{The overall clarity of this image is okay. The main subject, which is the boat, is relatively clear. The lighting is weak, making the overall image appear dark. The scenery in the distance is quite blurry, and the texture details are lost. The composition is good, but the quality of this image is poor}. Please help to identify the distortions of each region within the following types [blur, jitter, overexposure, low light, noise, no distortion].
\end{minipage}
\hfill
\begin{minipage}{0.49\linewidth}
    \includegraphics[width=0.8\textwidth]{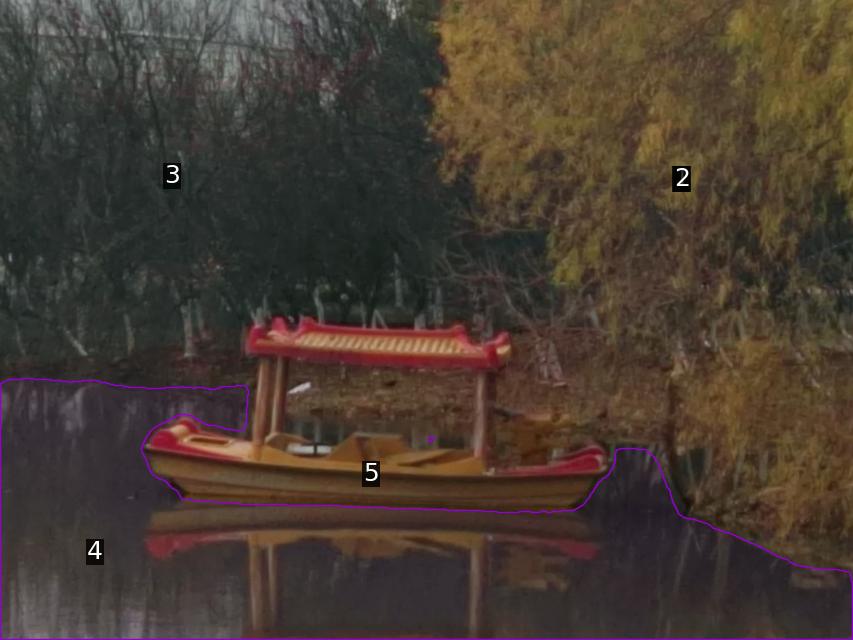}
    \vspace{1em}
\end{minipage}
    
\hdashrule[0.5ex]{\linewidth}{0.5pt}{0.1mm 0.5mm}

\textit{\textbf{\#GPT4V Response:}}

\begin{minipage}{0.59\linewidth}
\small
\setstretch{0.25}
\begin{lstlisting}[basicstyle=\ttfamily,columns=fullflexible,breaklines=true]
[
    {
        "2": "blur",
        "gpt4v iqa": "The scenery here is quite blurry, detail is lost."
    },
    {
        "3": "low light",
        "gpt4v iqa": "This area is dark and lacks adequate lighting."
    },
    {
        "4": "low light",
        "gpt4v iqa": "The image appears dark due to weak lighting."
    },
    {
        "5": "no distortion",
        "gpt4v iqa": "The main subject, the boat, appears relatively clear with no significant distortion."
    }
]
\end{lstlisting}
\end{minipage}
\hfill
\begin{minipage}{0.39\linewidth}
    \textit{\textbf{Manually generated mask} according to the text label from GPT4V and the corresponding SAM segmentation masks.}
    
    \includegraphics[width=0.9\textwidth]{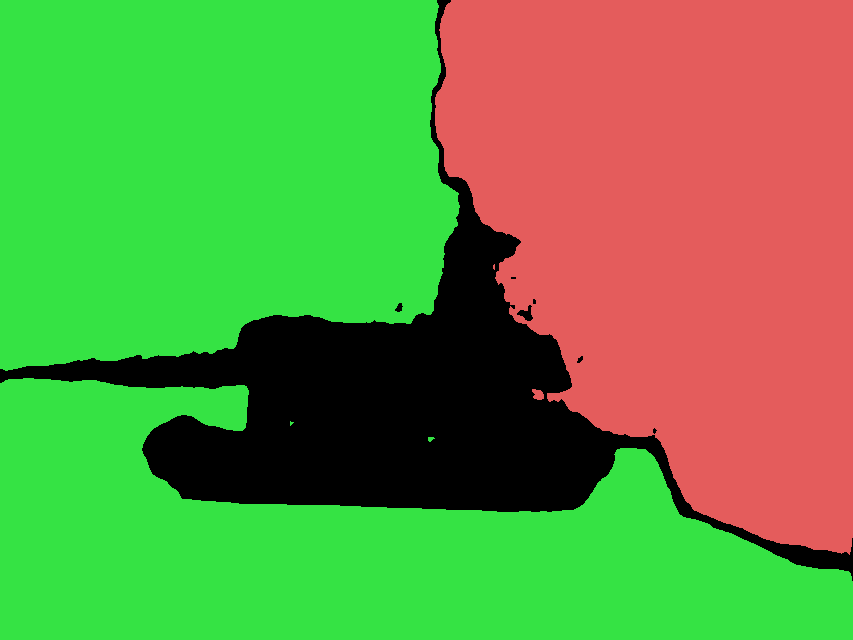}
\end{minipage}
\end{tcolorbox} 
\caption{Example of GPT4V annotation} \label{fig:gpt_anno}
\end{figure*}

The GPT4V annotation process is illustrated in \cref{fig:gpt_anno}. Within the system message, GPT4V is characterized as an effective IQA (Image Quality Assessment) assistant that recognizes five types of distortions, along with a \prompt{no distortion} category. Responses must adhere to the specified JSON format, where a short reasoning message is required to help verify the result. Users will provide quality prompts generated by the most recent Co-Instruct model\footnote{\url{https://huggingface.co/spaces/q-future/Co-Instruct}} using the designated prompt:
\begin{quote}
    \prompt{The input image: <|image|>. Describe and evaluate the quality of the image.}
\end{quote}
where \prompt{<|image|>} is the placeholder of image input. 
\Cref{fig:gpt_anno} presents an example of the GPT4V response, in which the correct answer is provided. We checked the response format and tried until it met our requirements. Finally, the corresponding regions were labelled with predicted distortions. We use shortest edit distance to find the best matching distortion types because the generated distortion words may not always exactly match the candidates.

\section{Training Details}

\subsection{Loss Functions}

As described in the main text, we use the following loss function to train the network:
\begin{equation}
    \mathcal{L} = \lambda_{txt} \mathcal{L}_{ce}(y_{txt}, \hat{y}_{txt}) + \lambda_{seg} \mathcal{L}_{seg}(y_{seg}, \hat{y}_{seg}).
\end{equation}
where $\mathcal{L}_{ce}$ is the auto-regressive cross-entropy loss, $\mathcal{L}_{seg}$ is the segmentation loss. We follow the same practice as \cite{lai2023lisa,ren2023pixellm} and use a combination of per-pixel binary cross-entropy loss and DICE loss for $\mathcal{L}_{seg}$ as following:
\begin{equation}
    \mathcal{L}_{seg} = \lambda_{bce} \textup{BCE}(y_{seg}, \hat{y}_{seg}) +  \lambda_{dice} \textup{DICE}(y_{seg}, \hat{y}_{seg}), 
\end{equation}
where the loss weights are set to $\lambda_{txt}=1.0$, $\lambda_{seg}=1.0$, $\lambda_{bce}=2.0$, $\lambda_{dice}=0.5$.

\subsection{Training Configurations}

We employed the DeepSpeed framework\footnote{\url{https://www.deepspeed.ai/}} to accelerate training and reduce memory requirements. The training was conducted using 4 NVIDIA 4090 GPUs. Given our modifications to the multi-modal projection block $\phi_v$, it was necessary to adhere to training protocols from LLaVA to align the visual and language representations. Consequently, the training process was structured into three phases:
\begin{enumerate}[leftmargin=*]
\item \textbf{Stage 1: Feature alignment between the vision encoder and the LLM.} During this phase, both the vision encoder and LLM were fixed. Training focused solely on the projector $\phi_v$ to align the vision and text representations.
\item \textbf{Stage 2: Visual instruction tuning.} This phase involved fine-tuning the model to enhance its capability to follow instructions, utilizing multi-modal instruction-following data.
\item \textbf{Stage 3: Mixture dataset tuning.} Once a robust base model was established, it was further finetuned to integrate visual quality grounding with other tasks.
\end{enumerate}
Hyperparameters for each stage are detailed in \cref{tab:hyper_param_s1} and \cref{tab:hyper_param_s23}. The entire training duration was approximately two days.

\begin{table}[t]
    \caption{Hyper-parameter configurations for \textbf{Stage 1}.}
    \label{tab:hyper_param_s1}
    \centering
    \renewcommand\arraystretch{1.3}
    \renewcommand\tabcolsep{12pt}
    \begin{tabular}{c|c}
    \hline
    Hyper-parameter config & Value \\ \hline
    \rowcolor{lightgray} Image encoder (frozen) & CLIP-L/14-336 \\
    \rowcolor{lightgray} LLM (frozen) & LLaVA-v1.5-7B\tablefootnote{\url{https://huggingface.co/liuhaotian/llava-v1.5-7b}}\\
    Input image size & $448\times448$ \\
    Layers used for $\phi_v$ & 7, 14, 23 \\ 
    \hdashline
    Optimizer & AdamW \\ 
    \rowcolor{pink!30} Learning rate & 5e-4 \\
    Weight decay & 0 \\
    $(\beta_1, \beta_2)$ & $(0.9, 0.95)$ \\
    Scheduler & WarmupCosineLR \\
    Warm up steps & 100 \\
    ZeRO stage (deepspeed) & 2 \\
    Precision & bfloat16 \\
    Batch size (with accumulation) & $2\times4\times10$ \\
    \rowcolor{pink!30} Training dataset & LAION-CC-SBU\tablefootnote{\url{https://huggingface.co/datasets/liuhaotian/LLaVA-Pretrain}} \\
    Total epochs & 1 
    \\ \hline
    \end{tabular}
\end{table}

\begin{table}[t]
    \caption{Hyper-parameter configurations for stage 2 and 3.}
    \label{tab:hyper_param_s23}
    \centering
    \renewcommand\arraystretch{1.3}
    \renewcommand\tabcolsep{8pt}
    \begin{tabular}{c|c}
    \hline
    Hyper-parameter config & Value \\ \hline
    \rowcolor{lightgray} Image encoder (frozen) & CLIP-L/14-336 \\
    LLM initialization &  Previous training stage \\
    Input image size & $448\times448$ \\
    Segmentation token number & 6 \\
    Layers used for $\phi_v$ & 7, 14, 23 \\ 
    LoRA rank & 64 \\
    LoRA $\alpha$ & 16 \\
    \hdashline
    Optimizer & AdamW \\ 
    \rowcolor{pink!30} Learning rate & 3e-4 \\
    Weight decay & 0 \\
    $(\beta_1, \beta_2)$ & $(0.9, 0.95)$ \\
    Scheduler & WarmupCosineLR \\
    Warm up steps & 100 \\
    ZeRO stage (deepspeed) & 2 \\
    Precision & bfloat16 \\
    Batch size (with accumulation) & $4\times4\times10$ \\
    \rowcolor{pink!30} Training dataset & \makecell[l]{\textbf{Stage 2}: Instruction dataset\tablefootnote{\url{https://huggingface.co/datasets/liuhaotian/LLaVA-Instruct-150K}} \\ \textbf{Stage 3}: Mixture dataset} \\ 
    Total samples & $80\times500\times20$ 
    \\ \hline
    \end{tabular}
\end{table}

\section{More Qualitative Results}

\begin{figure*}
    \centering
    \begin{subfigure}[b]{\textwidth}
        \centering
        \includegraphics[width=\linewidth]{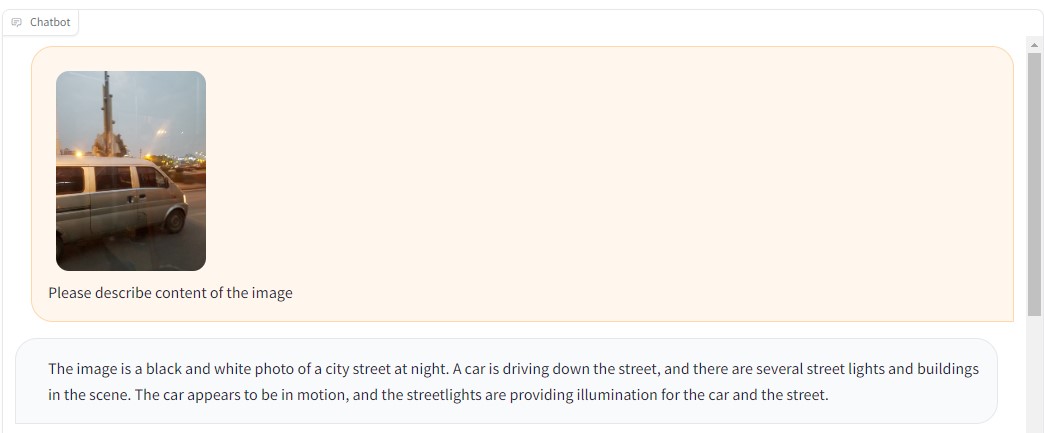} 
    \end{subfigure}
    \begin{subfigure}[b]{\textwidth}
        \centering
        \includegraphics[width=\linewidth]{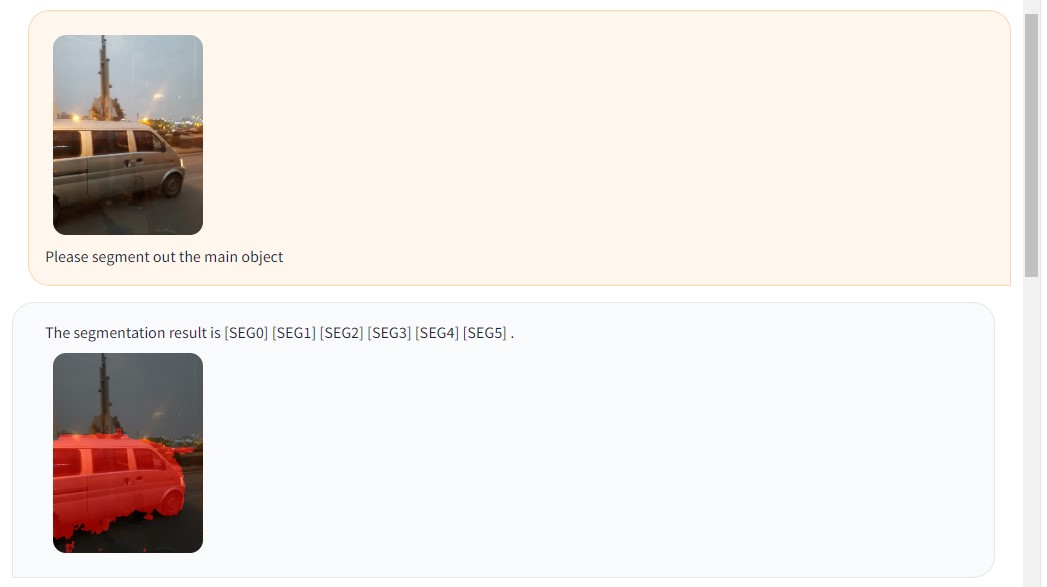} 
    \end{subfigure}
    \caption{Example of visual question answering and semantic segmentation.}
    \label{fig:eg01}
\end{figure*}

\begin{figure*}
    \begin{subfigure}[b]{\textwidth}
        \centering
        \includegraphics[width=\linewidth]{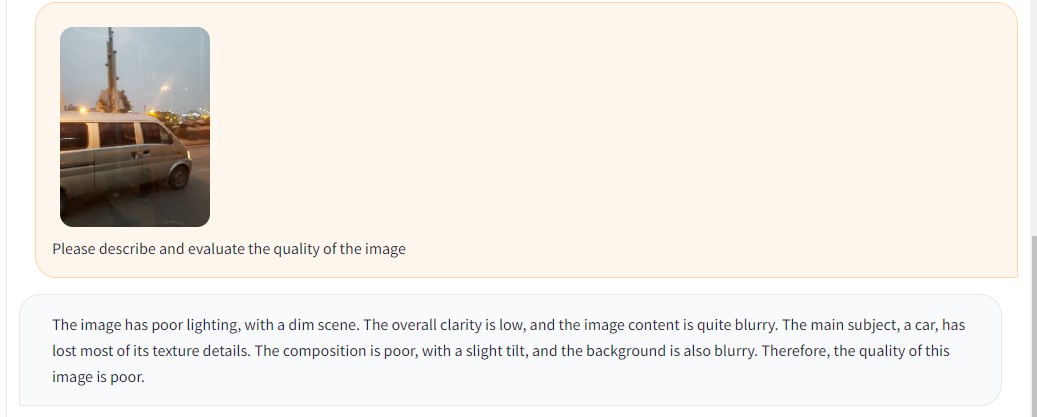} 
    \end{subfigure}
    \begin{subfigure}[b]{\textwidth}
        \centering
        \includegraphics[width=\linewidth]{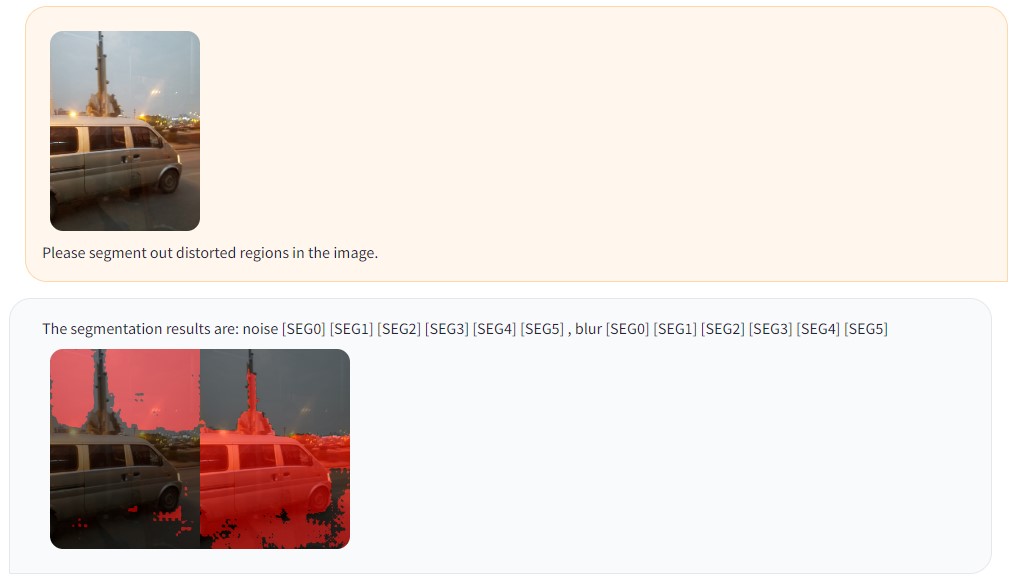} 
    \end{subfigure}
    \caption{Example of visual visual quality reasoning and visual quality grounding.}
    \label{fig:eg02}
\end{figure*}

Figures \ref{fig:eg01} and \ref{fig:eg02} demonstrate the versatile capabilities of our model.

\end{document}